# Talking the Talk Does Not Entail Walking the Walk: On the Limits of Large Language Models in Lexical Entailment Recognition

**Candida Maria Greco**
DIMES Department
University of Calabria
Rende, Italy

**Lucio La Cava**
DIMES Department
University of Calabria
Rende, Italy

**Andrea Tagarelli**
DIMES Department
University of Calabria
Rende, Italy

{candida.greco, lucio.lacava, tagarelli}@dimes.unical.it

## Abstract

Verbs form the backbone of language, providing the structure and meaning to sentences. Yet, their intricate semantic nuances pose a longstanding challenge. Understanding verb relations through the concept of *lexical entailment* is crucial for comprehending sentence meanings and grasping verb dynamics. This work investigates the capabilities of eight Large Language Models in recognizing lexical entailment relations among verbs through differently devised prompting strategies and zero-/few-shot settings over verb pairs from two lexical databases, namely WordNet and HyperLex. Our findings unveil that the models can tackle the lexical entailment recognition task with moderately good performance, although at varying degree of effectiveness and under different conditions. Also, utilizing few-shot prompting can enhance the models' performance. However, perfectly solving the task arises as an unmet challenge for all examined LLMs, which raises an emergence for further research developments on this topic.

## 1 Introduction

Verbs hold a central role in language, serving as the foundational and semantic framework for conveying sentence meaning. Understanding their semantic nuances has been a long challenge due to the considerable variability and relatively looser cohesion of verb meanings compared to nouns (Gentner and France, 1988). Exploring the relationships between verbs can be addressed by considering the concept of *lexical entailment* (Fellbaum, 1990; Geffet and Dagan, 2005). Analogous to logical entailment, which applies to propositions, lexical entailment describes the relationship between two verbs $v_1$ and $v_2$, wherein the sentence $\langle subject \rangle v_1$ logically entails the sentence $\langle subject \rangle v_2$, i.e., that one verb leads to another in the sentence.

Focusing on lexical entailment is paramount for several reasons. Verbs are crucial for expressing actions and relationships between entities, making it essential to properly capture their nuances. Grasping these relationships helps in deciphering sentence meanings and how verbs work together. Verbs often have polysemous and context-dependent meanings, and capturing entailment relations among verbs involves addressing challenges like aspect, modality, tense, and fine-grained semantic differences. Also, certain domains or specialized fields heavily rely on precise verb relations, such as in legal, scientific, or technical language.

The study of entailment has long been recognized as a critical endeavor in NLP and related fields, as it is fundamental to several tasks such as classification, summarization, question answering, and machine translation. Given their remarkable success in solving the aforementioned tasks, Large Language Models (LLMs) have indeed reshaped the landscape of language understanding (Chang et al., 2024; Min et al., 2024); nonetheless, to achieve even more sophisticated capabilities in interpreting human communication through verbal nuances, unveiling entailment recognition into these models represents a demanding challenge for their development. This has a number of motivations, which can be summarized as follows. Correctly inferring entailment relations between verbs is indeed essential for a LLM to be robust in understanding nuanced meanings and logical connections within sentences. By exploring how LLMs interpret and handle entailment among verbs, we can also gain insights into their decision-making processes, unveiling their strengths and weaknesses, and also contributing to model interpretability and refinement. As previously mentioned, handling verb entailment is crucial in various NLP tasks: on the one hand, improving models' comprehension of these relations can enhance the performance of these applications, but on the other hand, identifying and addressing biases or misinterpretations in entailment relations is also important in refining models to handle diverse linguistic contexts and minimize errors in real-world applications.

A related aspect, as recently noted in (Putra et al., 2023) for textual entailment in general, is an emergence for developing evaluation datasets and benchmarks for assessing the performance of models in entailment tasks. In this regard, we recognize the primary role played by *lexical databases*, which are meaningful resources for semantically exploring verb relations. In this work, we will focus on two widely used resources, namely WordNet (Miller et al., 1990) and HyperLex (Vulić et al., 2017).

Our study aims to unveil the actual capabilities of LLMs in recognizing lexical entailment relations. By focusing on instances of verb pairs corresponding to entailment relations provided by lexical databases, we conduct a thorough evaluation based on eight LLMs, with emphasis on Open LLMs. We define different prompting strategies with different context details, both under a zero-shot and a few-shot setting, for asking a LLM to answer about a question relating to the entailment between any two verbs. Our primary evaluation goal is to understand to what extent LLMs are able to recognize lexical entailment between verbs by measuring their compliance with well-grounded, manually curated linguistic resources.

In the remainder of this paper, we discuss related work in Sect. 2, the data sources and the selected LLMs in Sect. 3. Our defined methodology and experimental results are described in Sects. 4 and 5. Section 6 concludes the paper.

## 2 Related Work

We discuss recent studies involving LLMs, and more generally pre-trained language models (PLMs), to capture the meaning and connection between words. Sainz et al. (2023) address word sense disambiguation in terms of textual entailment to understand if BERT and RoBERTa can discriminate between different senses in a variety of domains. Tseng et al. (2023) train a mT5 model for generating Chinese word glosses, and raise the need for models to rely on semantic vectors.

Dealing with negation represents an additional challenge. Chen et al. (2023) assess negative commonsense knowledge of LLMs. Experiments carried out on Flan-T5 (Chung et al., 2024), GPT-3 (Brown et al., 2020), Codex (Chen et al., 2021), Instruct GPT (Ouyang et al., 2022), and ChatGPT[1] reveal behavior inconsistency among the LLMs. García-Ferrero et al. (2023) test the commonsense knowledge of open source LLMs (T5 (Raffel et al., 2020), Llama (Touvron et al., 2023a), Pythia (Biderman et al., 2023), Falcon (Almazrouei et al., 2023), Vicuna (Zheng et al., 2024)) using both affirmative and negative sentences using various types of relations and negations. Results have shown that the LLMs excel in classifying affirmative sentences but fail in dealing with negative ones.

More generally, PLMs and LLMs have been tested over various types of semantic relations. Lovón-Melgarejo et al. (2024) analyze the ability of BERT-based models and Sentence-Transformers to capture hierarchical semantic knowledge using WordNet-derived datasets. Oliveira (2023) apply BERT to capture synonyms, antonyms, hypernyms, and hyponyms in the Portuguese language. Hypernymy is also a focus of the study in (Liao et al., 2023), which builds a dataset on the WordNet hypernyms and test several PLMs and LLMs on hypernymy discovery, observing a consistent underperformance of LLMs when tasked with abstract concepts. Bai et al. (2022) assign words with a common WordNet hypernym into the same class, and train PLMs by gradually transitioning from predicting the class to predicting the token through a curriculum learning strategy. Also, Tikhomirov and Loukachevitch (2024) evaluate the use of LLMs with various prompts for hypernym prediction.

Other studies investigated how to leverage the relationships and definitions provided by WordNet to enhance the representation through generated embeddings and data augmentation (Loureiro and Jorge, 2019; Perçin et al., 2022).

Our work uniquely provides an analysis of how currently used LLMs can recognize verb entailment relations, in contrast with works with broader scopes like Lovón-Melgarejo et al. (2024) and Tikhomirov and Loukachevitch (2024). Our focus on verbs also differs from studies such as Sainz et al. (2023), which consider an organization of relations into domain-specific and high-level concepts. Unlike Tseng et al. (2023), we do not focus on attempting to determine a specific sense for a sentence, instead our aim is to understand how LLMs answer to specific queries about semantic relations. Our methodology differs from Chen et al. (2023) and García-Ferrero et al. (2023), since they consider negative relations and not necessarily on verbs. Moreover, we focus on a representative set of recently developed open and commercially-licensed LLMs, while García-Ferrero et al. (2023) consider open models only and Chen et al. (2023) focus on models earlier than ChatGPT. Oliveira (2023) is limited to Portuguese and uses BERT to determine relations of hyponymy, hypernymy, synonymy, and antonymy. Unlike Liao et al. (2023), which train projection layers to learn WordNet hypernym relations, we approach the problem through a probing approach. Compared to Bai et al. (2022), which focuses on reducing model perplexity, our interest is understanding how models specifically handle verb entailment relations.

[1] https://openai.com/blog/chatgpt

## 3 Resources used in this study

### 3.1 Data

We resorted to two widely recognized and openly accessible lexical resources, which provided us with the means to address our research questions regarding the LLM awareness of verb entailment.

**WordNet** (Miller et al., 1990) is a well-known large lexical database of English, providing features for different uses as online dictionary, thesaurus, and lexical ontology. WordNet stores terms into lexical source files by syntactic categories, i.e., nouns, verbs, adjectives, adverbs, which are grouped into sets of cognitive synonyms, called *synsets*, each expressing a distinct (lexicalized) concept. Focusing on verbs, they are categorized according to semantic fields corresponding to 15 lexicographers' files: motion, perception, communication, competition, change, cognitive, consumption, interaction, creation, emotion, possession, body care, social behavior, weather, stative functions. Note that verbs can appear in various forms, including monadic, phrasal verbs, and compound verbs.

Entailment relations between verb synsets can in principle be distinguished as hyponyms or troponyms (and their reverse form, i.e., hypernyms), antonyms, and (other kinds of) entailments. In particular, *troponymy* is a special case of entailment, since a verb $v_1$ is a troponym of verb $v_2$ if the activity (corresponding to) $v_1$ is doing $v_2$ in some manner; moreover, antonyms can also be troponyms (e.g., fail/succeed entails try, forget entails know). In practice, however, WordNet verb entailments can be accessed via either *hyponym* or *entailment set function*, as follows: if $v_1$ entails $v_2$, then in WordNet either $v_1$ belongs to the set of $v_2$'s hyponyms or $v_2$ belongs to the set of $v_1$'s entailments.

| Model | Reference | # Params | Owner |
|---|---|---|---|
| gpt-3.5-turbo | (Brown et al., 2020) | n.a. | OpenAI |
| Meta-Llama-3-8B-Instruct | (Dubey et al., 2024) | 8B | Meta |
| Llama-2-7b-chat-hf | (Touvron et al., 2023b) | 7B | Meta |
| Mistral-7B-Instruct-v0.1 | (Jiang et al., 2023) | 7B | Mistral |
| falcon-7b-instruct | (Almazrouei et al., 2023) | 7B | TII |
| vicuna-7b-v1.5 | (Zheng et al., 2024) | 7B | LMSYS |
| neural-chat-7b-v3-2 | n.a. | 7B | Intel |
| gemma-1.1-7b-it | (Team et al., 2024) | 7B | Google |

Table 1: Summary of the LLMs used in this work. Model names but GPT refer to HuggingFace Hub tags.

**HyperLex** (Vulić et al., 2017) was built as a gold standard resource for measuring and evaluating how well semantic models capture *graded* or soft lexical entailment. To this aim, HyperLex data contain 2616 word pairs, of which 453 are verb pairs, associated with asymmetric scores on a scale 0-6 that were annotated by humans according to the question "To what degree is X a type of Y?".

WordNet and HyperLex have been widely recognized as a valuable support for entailment tasks, both as sources of knowledge and benchmarks. An example of this complementary contribution is the study in (Renner et al., 2023), which showcases the utility of WordNet in supporting *graded lexical entailment* (GLE) tasks, i.e. assigning a degree of the lexical entailment relation between two concepts, demonstrating how leveraging hierarchical synset structures can improve performance, as also assessed by considering the HyperLex dataset as a benchmark in experimental evaluation.

### 3.2 LLMs

Our study involves a representative selection of LLMs, which reflect various baseline architectures. Specifically, we use GPT-3.5 (Brown et al., 2020) through the official OpenAI APIs, along with some of the most prominent *Open* LLMs (i.e., *open-source* or *open-weights*), namely Llama-3 (Dubey et al., 2024) in its 8B-parameter version, the 7B-parameter versions of Llama-2 (Touvron et al., 2023b), Mistral (Jiang et al., 2023), Falcon (Almazrouei et al., 2023), Vicuna (Zheng et al., 2024), Gemma (Team et al., 2024), and Intel NeuralChat.

Table 1 summarizes essential information about the models involved in this study. For each model we provide the following: (i) the specific instance name; (ii) the associated publication, if available; (iii) the size of the model expressed in billions of parameters, if available; (iv) the company which trained the model; (v) the reference to the implementation used.

# 4 Methodology

## 4.1 Evaluation data

**WordNet.** To access WordNet verbs and entailments, we resort to the implementations provided by the NLTK[2] library. According to the logical organization in WordNet, verb entailments can be retrieved through two methods, namely *hyponyms()* and *entailments()*, where the former provides access to troponymies, and the latter includes the other kinds of entailment relations as described in (Fellbaum, 1990). Given a target verb synset, either method returns its associated set of hyponym/entailment synsets. To be consistent with the WordNet verb hierarchy which considers both direct and indirect hyponyms, we used the *hyponyms()* method recursively to get all hyponyms of a given synset.

To reduce the synset relations to lemma relations, a further step is to expand each pair of synsets as a set resulting from the Cartesian product of their respective lemma sets. Note also that each synset is provided with its definition (gloss), which transfers to each of its constituting lemmas.

By performing the above steps, we retrieved 114,490 lemma pairs based on *hyponyms()* and 2,352 lemma pairs based on *entailments()*. Therefore, the total of WordNet verb pairs that are **relevant** to the task under study is 114,490 + 2,352 = 116,842. We then finally built our WordNet evaluation dataset by selecting as many verb-pairs that are **not relevant** to the task, by randomly picking 50% of them by rewiring the relevant pairs to obtain non-relevant pairs, and the other 50% from the complement set of WordNet verbs that are not involved in any type of entailment relation.

**HyperLex.** As mentioned in Sect. 3.1, HyperLex provides 380 verb pairs corresponding to different types of entailment relations: hyponymy, hypernymy, co-hyponymy, synonymy and antonymy. Among these, we notice that 169 verb pairs also appear in the set of WordNet pairs, six of which are labeled as synonyms and other six as hypernyms, while the remaining ones in HyperLex either are not present in entailment relations in WordNet or they are provided in reverse order; in particular, HyperLex has 71 reciprocal verb-pairs, of which 27 are shared with WordNet. Likewise in WordNet, we couple the set of 380 pairs (**relevant**) with as many verb pairs as **not relevant**, obtained by rewiring the relevant ones.

[2] https://www.nltk.org/

## 4.2 Prompts

We crafted three prompt schemes, which correspond to different types of instructions for the models. Also, for each prompt, we devised both a *zero-shot* scenario and a *few-shot scenario*, whereby we enhance the prompt with contextual examples.

We tailored the prompt setting depending on the underlying resource. When using verbs from WordNet, we augment the prompt by incorporating into it the definition of each verb lemma used in the prompt (i.e., the gloss of the corresponding synset from which the lemma is derived); we will use symbol $def()$ to denote a function returning the definition of a verb lemma. By contrast, for HyperLex, we do not augment the prompts since verb definitions are not originally provided within this resource. These two settings will enable us to investigate the impact of the presence/absence of verb definitions on the lexical entailment recognition capabiliies of the examined LLMs.

Next we introduce our defined prompts (we shall specify by "[...]" the optionality of verb definitions). We begin with the *zero-shot* prompts we used as the first step of our experimental evaluation.

**Direct Entailment.** Our first type of prompt, dubbed as Direct, is devised to test the ability of a model to recognize (any) entailment relation between two verbs:

**Direct Prompt**

*Given the verb $v_1$ [defined as $def(v_1)$] and the verb $v_2$ [defined as $def(v_2)$], what is the verb that entails the other?*
*Answer must be either of the form "X entails Y" or "there is no entailment".*

It should be noted that, through the Direct prompt, the model is required implicitly to first recognize the existence of entailment, and in this case, to decide which verb is the entailing verb, and which verb is the entailed verb.

**Indirect Entailment.** The Direct prompt requires the model to rely on its knowledge of the meaning of the word "entail", thus explicitly framing an entailment recognition task. By contrast, our second type of prompt omits the use of "entail" and instead provides the model with a relational function that expresses a definition of entailment. We refer to this prompt type as Indirect:

**Indirect Prompt**

*Relation F states that given two verbs X and Y, X and Y satisfy F if and only if when doing Y you are also doing X.*

*Given the verb $v_1$ [defined as $def(v_1)$] and the verb $v_2$ [defined as $def(v_2)$], what is X and what is Y?*
*Answer must be either a pair (X,Y) or "relation F cannot be satisfied".*

Like for the Direct prompt, to answer the Indirect prompt the model needs to decide the correct order between entailing and entailed verbs, otherwise to recognize the case whereby no entailment relation holds for the two input verbs.

**Reverse Entailment.** Our third type of prompt is designed to allow us to determine whether the model can recognize the entailment relation in the presence of negation. That is, for each verb pair $\langle v_1, v_2 \rangle$ such that $v_1$ entails $v_2$, we ask the model whether it holds true that "not $v_2$ entails not $v_1$". We refer to this prompt type as Reverse:

**Reverse Prompt**

*Given the verb $v_1$ [defined as $def(v_1)$] and the verb $v_2$ [defined as $def(v_2)$], answer YES if 'not $v_2$' entails 'not $v_1$,' NO otherwise.*

Note that, despite the binary nature of its required response, the Reverse prompt poses a different challenge than the other two types in that the model is required to "reason" about a reverse entailment based on negation.

**Few-shot settings.** Our previously defined prompts are also used to perform in-context learning based on contextual examples. To this purpose, we define two few-shot scenarios, hereinafter referred to as *HyperLex-based few-shot examples* (HyperLex-FS) and as *Fellbaum-based few-shot examples* (Fellbaum-FS). In the former case, we select and introduce in a prompt four verb pairs that correspond to different difficulty levels based on the scores assigned by HyperLex to the verb pair. In the latter case, we introduce in a prompt one example for each of the four types of entailment relations described in (Fellbaum, 1990), namely *troponymy co-extensiveness*, *troponymy proper inclusion*, *backward presupposition* and *cause*. In both cases, the set of examples are fixed for all test verb-pairs, except when an example pair coincides with the test verb-pair (in which case, the example pair is replaced with another equivalent according to the particular strategy used).

Full details about our defined zero-shot prompts, few-shot prompts, and selection of the contextual examples are reported in Appendices A–C.

### 4.3 Model settings and deployment

As we leverage models that are aligned to human preferences, being them of *instruct* or *chat* type, we specified a "system prompt" to declare the model's *role*. To this purpose, we set each model to be *"a linguistic expert who responds to user questions in a concise way."*

A key aspect of our model deployment is the use of the *Guidance* library[3] through its `select` method to achieve *constrained generation* (Liu et al., 2023). This forces an LLM to produce *structured* outputs that adhere to the shape and contents required by our defined prompts (Sect. 4.2). This way, not only the models' outcomes will be strictly pertinent to the admissible or valid responses, but also there is no constraint to set on the maximum number of tokens to generate. The model temperature was set to the minimum allowed value for any particular model, which is determined by the Guidance framework through its `select` method to achieve constrained generation. In reality, the temperature value is a number that is kept below 0.01, so as to avoid division by zero in the decoding probability computation. Likewise, we refrained from altering the *top_p* and *top_k* parameters from their default values of 50 and 1, respectively.

We carried out our experiments locally by deploying the models through the open-source *text-generation-webui* framework,[4] using a 8x NVIDIA A30 GPU server with 24 GB of RAM each, 764 GB of system RAM, a Double Intel Xeon Gold 6248R with a total of 96 cores, and Ubuntu Linux 20.04.6 LTS as operating system.

### 4.4 Evaluation criteria

To validate the quality of the responses generated by the models, we resorted to two approaches: the first one based on standard statistical assessment criteria for a classification task, and the second one based on a *model's self-evaluation*.

The former group includes *accuracy*, *precision*, *recall*, and $F1$-*score* calculated by comparing a model's responses to the available ground-truth, for all models and prompt settings. Such measures were calculated upon the confusion matrix definition corresponding to each of three prompt types, as summarized in Appendix D.

[3] https://github.com/guidance-ai/guidance
[4] https://github.com/oobabooga/text-generation-webui

The model's self-evaluation (Xiong et al., 2024) was carried out by asking the model to provide its response with a *confidence* score in [1..10], where higher scores correspond to higher degree of certainty by the model in deciding about entailment between two verbs. In the results, we shall report ***r-conf*** resp. ***nr-conf*** to denote the confidence of a model averaged over all relevant pairs resp. not-relevant pairs.

## 5 Results

We organize the result presentation into two subsections, which correspond to evaluation on WordNet data, resp. HyperLex data. We shall first discuss results obtained in a zero-shot scenario, then those in the few-shot scenarios. In the tables, bold resp. underlined values will correspond to the highest resp. second-highest scores per column.

### 5.1 Evaluation on WordNet data

**Zero-Shot Prompting.** Table 2 reports the results obtained under the zero-shot setting on WordNet verb pairs. Let us first consider results corresponding to the Direct prompting. NeuralChat and Gemma emerge as the best-performing models, although according to different assessment criteria: NeuralChat excels in precision (0.95) and accuracy (0.62), followed by Llama-3, whereas Gemma outperforms the others in recall (0.89) and F1-score (0.62), with Vicuna as the second best. This indicates relatively more robustness by models like NeuralChat and Llama-3 against false positives and true negatives, while other models like Gemma and Vicuna behave better in terms of false negatives.

The above scenario changes when prompting the models through Indirect, i.e., with an explicit definition of entailment without using any cue word. The observed change regards not just the measurement scores (which correspond to higher F1 on average) but also the best performing models, which are now Llama-2, Mistral and Vicuna.

The Reverse prompting corresponds to a different scenario, whereby GPT-3.5 emerges as the most effective model according to all criteria, followed by Llama-3, Vicuna, and NeuralChat. Despite the negation-based reversed form of the entailment relation to recognize, the good behavior of GPT-3.5 under Reverse contrasts with the disappointing results obtained under the other two prompts, which would suggest that GPT-3.5 is more suited to deal with the task at hand when prompted to provide a yes/no answer.

Interesting remarks can also be drawn from a comparison of the behaviors of the models w.r.t. their architectural commonalities. For instance, Vicuna can behave generally better than the Llama models in terms of recall and F1-score, despite sharing most of the same architecture with them, which might be explained as an effect of the fine-tuning of Vicuna that was carried out based on user-shared ChatGPT conversations collected from ShareGPT.com. Likewise, when using the Direct and Reverse prompts, NeuralChat outperforms Mistral despite being based on it, which can be attributed to the fine-tuning process using the higher-quality *SlimOrca* dataset (Mukherjee et al., 2023), and to a *Direct Preference Optimization* (DPO) phase,[5] designed to better align the model with human preferences.

Another interesting remark regards the confidence expressed by the models in answering the prompts, which is consistently very high, or even equal to the maximum (10.0) for Falcon and NeuralChat. One exception is represented by Gemma which, despite being a reliable model under all promptings, consistently associates its answers with confidence 2, as it was inherently extremely cautious, or uncertain, on the task.

**Prompting with HyperLex-FS.** Our results based on the HyperLex-FS strategy (Table 3-top) show that some models can better recognize verb entailments when supported by contextual examples. This holds especially for GPT-3.5 and Llama-3 under Direct and Indirect promptings (with percentage increase in F1 of 92% for Llama-3 under Direct and of 277% for GPT-3.5 under Indirect), but also Falcon, Vicuna and Mistral benefit from the HyperLex-FS strategy.

By contrast, such models as Gemma under Indirect and Reverse, Mistral under Direct and Reverse, and NeuralChat under Direct, are unexpectedly damaged by the use of the HyperLex-FS prompting, as it turns out to break their ability to properly answer. This might be ascribed to the increased complexity of the prompt, rather than its length, which however was ensured to be within the models' limits of maximum token length.

[5]huggingface.co/datasets/Intel/orca_dpo_pairs

| | Direct | | | | | | Indirect | | | | | | Reverse | | | | | |
|---|---|---|---|---|---|---|---|---|---|---|---|---|---|---|---|---|---|---|
| | $A$ | $P$ | $R$ | $F1$ | r-conf | nr-conf | $A$ | $P$ | $R$ | $F1$ | r-conf | nr-conf | $A$ | $P$ | $R$ | $F1$ | r-conf | nr-conf |
| gpt-3.5 | 0.144 | 0.151 | 0.154 | 0.152 | 8.93 | 8.12 | 0.082 | 0.141 | 0.164 | 0.151 | 8.89 | 9.36 | **0.629** | 0.577 | 0.965 | **0.722** | 9.29 | 9.14 |
| llama-3 | 0.571 | 0.862 | 0.169 | 0.283 | 9.94 | 9.95 | 0.334 | 0.392 | 0.605 | 0.476 | 9.07 | 8.65 | 0.605 | 0.579 | 0.767 | 0.660 | 9.74 | 9.67 |
| llama-2 | 0.422 | 0.363 | 0.205 | 0.262 | 8.00 | 8.00 | **0.670** | **0.623** | 0.860 | **0.723** | 7.98 | 8.00 | 0.539 | 0.599 | 0.236 | 0.339 | 7.99 | 7.99 |
| mistral | 0.536 | 0.604 | 0.210 | 0.311 | 8.97 | 9.13 | 0.470 | 0.485 | **0.936** | 0.639 | 9.00 | 9.00 | *0.500* | *NaN* | *0.000* | *NaN* | *9.00* | *9.00* |
| falcon | 0.352 | 0.263 | 0.165 | 0.202 | **10.00** | **10.00** | 0.419 | 0.456 | 0.839 | 0.591 | **10.00** | **10.00** | 0.527 | 0.515 | 0.909 | 0.658 | **10.00** | **10.00** |
| vicuna | 0.459 | 0.463 | 0.515 | 0.488 | 9.00 | 9.00 | 0.478 | 0.488 | 0.920 | 0.638 | 9.00 | 9.00 | 0.500 | 0.500 | **1.000** | 0.667 | 9.00 | 9.00 |
| neural-chat | **0.622** | **0.946** | 0.259 | 0.407 | **10.00** | **10.00** | 0.360 | 0.418 | 0.718 | 0.529 | **10.00** | **10.00** | 0.616 | **0.807** | 0.305 | 0.442 | **10.00** | **10.00** |
| gemma | 0.450 | 0.473 | **0.881** | **0.616** | 2.00 | 2.00 | 0.421 | 0.456 | 0.820 | 0.586 | 2.00 | 2.00 | 0.558 | 0.551 | 0.629 | 0.588 | 2.00 | 2.00 |

Table 2: Zero-shot prompting results on WordNet verb pairs.

| | Direct | | | | | | Indirect | | | | | | Reverse | | | | | |
|---|---|---|---|---|---|---|---|---|---|---|---|---|---|---|---|---|---|---|
| | $A$ | $P$ | $R$ | $F1$ | r-conf | nr-conf | $A$ | $P$ | $R$ | $F1$ | r-conf | nr-conf | $A$ | $P$ | $R$ | $F1$ | r-conf | nr-conf |
| gpt-3.5 | 0.318 | 0.240 | 0.169 | 0.198 | 8.13 | 7.37 | 0.398 | 0.443 | 0.796 | 0.569 | 7.66 | 8.18 | **0.664** | 0.613 | **0.890** | **0.726** | 7.58 | 7.60 |
| llama-3 | **0.660** | **0.831** | 0.402 | 0.542 | 9.71 | 9.30 | 0.634 | 0.606 | 0.767 | 0.677 | 9.77 | 9.27 | 0.617 | 0.958 | 0.244 | 0.389 | 8.94 | 8.71 |
| llama-2 | 0.358 | 0.256 | 0.149 | 0.188 | 8.01 | 8.00 | 0.048 | 0.087 | 0.095 | 0.091 | 8.00 | 8.00 | 0.539 | 0.602 | 0.231 | 0.333 | 8.00 | 8.00 |
| mistral | *0.500* | *NaN* | *0.000* | *NaN* | *8.75* | *8.86* | **0.722** | **0.749** | 0.668 | **0.706** | 9.00 | 9.00 | *0.500* | *NaN* | *0.000* | *NaN* | 6.78 | 7.20 |
| falcon | 0.444 | 0.436 | 0.385 | 0.409 | **10.00** | **10.00** | 0.447 | 0.472 | **0.893** | 0.617 | **10.00** | **10.00** | 0.557 | 0.565 | 0.501 | 0.531 | 9.90 | 9.91 |
| vicuna | 0.572 | 0.574 | **0.557** | **0.566** | 8.96 | 8.95 | 0.357 | 0.416 | 0.709 | 0.524 | 9.00 | 9.00 | 0.605 | 0.743 | 0.320 | 0.447 | 9.00 | 9.00 |
| neural-chat | *0.500* | *NaN* | *0.000* | *NaN* | *10.00* | *10.00* | 0.330 | 0.389 | 0.594 | 0.470 | **10.00** | **10.00** | 0.583 | **0.972** | 0.172 | 0.292 | **10.00** | **9.99** |
| gemma | 0.353 | 0.341 | 0.316 | 0.328 | 2.00 | 2.00 | *0.500* | *NaN* | *0.000* | *NaN* | *2.00* | *2.00* | *0.500* | *NaN* | *0.000* | *NaN* | *2.00* | *2.00* |

| | Direct | | | | | | Indirect | | | | | | Reverse | | | | | |
|---|---|---|---|---|---|---|---|---|---|---|---|---|---|---|---|---|---|---|
| | $A$ | $P$ | $R$ | $F1$ | r-conf | nr-conf | $A$ | $P$ | $R$ | $F1$ | r-conf | nr-conf | $A$ | $P$ | $R$ | $F1$ | r-conf | nr-conf |
| gpt-3.5 | 0.426 | 0.358 | 0.187 | 0.245 | 8.03 | 7.33 | 0.416 | 0.454 | 0.832 | 0.587 | 7.52 | 8.09 | **0.659** | 0.626 | 0.789 | **0.698** | 7.62 | 7.62 |
| llama-3 | **0.661** | **0.617** | **0.852** | **0.715** | 9.29 | 8.54 | 0.508 | 0.505 | 0.761 | 0.607 | 9.87 | 9.57 | 0.617 | 0.942 | 0.251 | 0.396 | 9.23 | 9.03 |
| llama-2 | 0.290 | 0.271 | 0.248 | 0.259 | 8.00 | 8.00 | 0.120 | 0.194 | 0.240 | 0.215 | 8.00 | 8.00 | 0.526 | 0.523 | 0.585 | 0.552 | 8.00 | 8.00 |
| mistral | *0.500* | *NaN* | *0.000* | *NaN* | *8.85* | *8.87* | **0.658** | **0.715** | 0.525 | 0.605 | 9.00 | 9.00 | *0.500* | *NaN* | *0.000* | *NaN* | *7.07* | *7.38* |
| falcon | 0.442 | 0.436 | 0.392 | 0.413 | **10.00** | **10.00** | 0.461 | 0.480 | **0.922** | **0.631** | **10.00** | **10.00** | 0.522 | 0.512 | **0.947** | 0.665 | 9.94 | 9.96 |
| vicuna | 0.597 | 0.598 | 0.593 | 0.595 | 8.96 | 8.95 | 0.458 | 0.473 | 0.734 | 0.576 | 8.99 | 9.00 | 0.592 | 0.738 | 0.286 | 0.412 | 9.00 | 9.00 |
| neural-chat | *0.500* | *NaN* | *0.000* | *NaN* | *10.00* | *10.00* | 0.414 | 0.452 | 0.813 | 0.581 | **10.00** | **10.00** | 0.577 | **0.964** | 0.160 | 0.274 | **10.00** | **9.99** |
| gemma | 0.353 | 0.341 | 0.316 | 0.328 | 2.00 | 2.00 | *0.500* | *NaN* | *0.000* | *NaN* | *2.00* | *2.00* | *0.500* | *NaN* | *0.000* | *NaN* | *2.00* | *2.00* |

Table 3: Few-shot prompting results on WordNet verb pairs: (top) HyperLex-FS, (bottom) Fellbaum-FS.

The higher complexity of the HyperLex-FS-based prompting w.r.t. the zero-shot scenario appears not to impact on the models' confidence, which remains equally high, with Falcon and (partly) NeuralChat confirming to be perfectly confident models.

**Prompting with Fellbaum-FS.** Let us now consider results corresponding to the Fellbaum-FS-based prompting which exploits examples of entailment relations in the Fellbaum taxonomy. Results are shown in Table 3-bottom.

One major remark is that the Fellbaum-FS strategy allows some models to further improve their ability of recognizing entailments. Compared to HyperLex-FS, significant improvements occur for the Llama models and Falcon, but also for GPT-3.5 and Vicuna under Direct and Indirect.

The Fellbaum-FS strategy turns out to be the best setting in absolute for Llama-3, GPT-3.5 and Vicuna under Direct, for Falcon, GPT-3.5 and NeuralChat under Indirect, and partly for Falcon, GPT-3.5 and Llama-3 under Reverse. Also, as already observed for the HyperLex-FS strategy, the performance of certain models is inhibited by the incorporation of usage examples in the prompt. Yet, the models' confidence values remain very similar to those observed for the HyperLex-FS strategy.

**Summary.** Our evaluation of lexical entailment recognition over WordNet verb pairs has unveiled that the examined LLMs can deal with the task showing moderate effectiveness on average. While no absolute winner emerges among the models, they tend to better understand Indirect than Direct in the zero-shot prompting, although in general the models can deal with all three types with comparable results on average. Interestingly, the skills of some models, especially under Direct and Indirect, tend to be improved by using HyperLex-FS and especially Fellbaum-FS. Llama-3 and GPT-3.5 reveal to be the models that mostly benefit from the few-shot prompting strategies. In terms of self-confidence, all models but Gemma exhibit high values, regardless of being successful in solving the task or degenerating to a constant answer, like it can happen for Mistral, Gemma and NeuralChat under certain conditions. Further details on the impact of WordNet relation types and the distribution of lexname categories are discussed in Appendix E.

| | Direct | | | | | | Indirect | | | | | | Reverse | | | | | |
|---|---|---|---|---|---|---|---|---|---|---|---|---|---|---|---|---|---|---|
| | $A$ | $P$ | $R$ | $F1$ | r-conf | nr-conf | $A$ | $P$ | $R$ | $F1$ | r-conf | nr-conf | $A$ | $P$ | $R$ | $F1$ | r-conf | nr-conf |
| gpt-3.5 | 0.197 | 0.089 | 0.066 | 0.076 | 8.520 | 8.043 | 0.024 | 0.045 | 0.047 | 0.046 | 8.556 | 8.664 | 0.514 | 0.507 | 0.995 | **0.672** | 8.870 | 8.553 |
| llama-3 | 0.405 | 0.409 | 0.426 | 0.418 | 9.075 | 8.476 | 0.271 | 0.352 | 0.542 | 0.427 | 9.325 | 8.996 | 0.625 | 0.673 | 0.487 | 0.565 | 9.648 | 9.425 |
| llama-2 | 0.355 | 0.353 | 0.347 | 0.350 | 8.000 | 8.000 | **0.504** | **0.636** | 0.018 | 0.036 | 6.449 | 6.281 | 0.533 | 0.727 | 0.105 | 0.184 | 7.968 | 7.961 |
| mistral | 0.533 | 0.562 | 0.300 | 0.391 | 9.351 | 9.228 | 0.479 | 0.489 | **0.958** | **0.648** | 9.000 | 9.008 | 0.500 | NaN | 0.000 | NaN | 7.366 | 7.413 |
| falcon | 0.391 | 0.294 | 0.155 | 0.203 | **10.000** | **10.000** | 0.375 | 0.429 | 0.750 | 0.545 | **10.000** | **10.000** | 0.503 | 0.502 | 0.668 | 0.573 | **10.000** | **10.000** |
| vicuna | 0.346 | 0.403 | 0.639 | 0.494 | 9.000 | 9.000 | 0.467 | 0.483 | 0.934 | 0.637 | 9.000 | 9.000 | 0.500 | 0.500 | **1.000** | 0.667 | 9.000 | 9.000 |
| neural-chat | **0.579** | **0.885** | 0.182 | 0.301 | **10.000** | **10.000** | 0.345 | 0.408 | 0.689 | 0.513 | **10.000** | **10.000** | **0.689** | **0.919** | 0.416 | 0.572 | **10.000** | **10.000** |
| gemma | 0.466 | 0.482 | **0.932** | **0.636** | 2.000 | 2.000 | 0.159 | 0.239 | 0.313 | 0.271 | 2.000 | 2.000 | 0.555 | 0.672 | 0.216 | 0.327 | 2.000 | 2.000 |

Table 4: Zero-shot prompting results on HyperLex verb pairs.

| | Direct | | | | | | Indirect | | | | | | Reverse | | | | | |
|---|---|---|---|---|---|---|---|---|---|---|---|---|---|---|---|---|---|---|
| | $A$ | $P$ | $R$ | $F1$ | r-conf | nr-conf | $A$ | $P$ | $R$ | $F1$ | r-conf | nr-conf | $A$ | $P$ | $R$ | $F1$ | r-conf | nr-conf |
| gpt-3.5 | 0.329 | 0.207 | 0.121 | 0.153 | 8.504 | 7.839 | 0.300 | 0.375 | 0.600 | 0.462 | 8.474 | 8.936 | 0.501 | 0.501 | 0.929 | 0.651 | 7.898 | 7.799 |
| llama-3 | **0.711** | **0.666** | **0.845** | **0.745** | 8.896 | 8.286 | **0.557** | **0.537** | 0.826 | **0.651** | **10.000** | 9.947 | 0.551 | 0.915 | 0.113 | 0.201 | 8.048 | 8.035 |
| llama-2 | 0.504 | 0.516 | 0.126 | 0.203 | 8.000 | 8.000 | 0.042 | 0.078 | 0.084 | 0.081 | 8.000 | 8.000 | 0.499 | 0.499 | **0.997** | **0.665** | 8.000 | 8.000 |
| mistral | 0.500 | NaN | 0.000 | NaN | 9.000 | 9.000 | 0.321 | 0.361 | 0.463 | 0.406 | 9.000 | 9.000 | 0.500 | NaN | 0.000 | NaN | 7.755 | 7.903 |
| falcon | 0.414 | 0.406 | 0.371 | 0.388 | **10.000** | **10.000** | 0.451 | 0.474 | **0.903** | 0.622 | **10.000** | **10.000** | 0.501 | **1.000** | 0.003 | 0.005 | 9.961 | 9.966 |
| vicuna | 0.495 | 0.496 | 0.584 | 0.536 | 7.862 | 8.448 | 0.204 | 0.290 | 0.408 | 0.339 | 9.000 | 9.000 | **0.593** | 0.771 | 0.266 | 0.395 | 9.000 | 9.000 |
| neural-chat | 0.500 | NaN | 0.000 | NaN | **10.000** | **10.000** | 0.229 | 0.195 | 0.174 | 0.184 | **10.000** | **10.000** | 0.558 | 0.907 | 0.129 | 0.226 | **10.000** | **10.000** |
| gemma | 0.321 | 0.321 | 0.321 | 0.321 | 2.000 | 2.000 | 0.500 | NaN | 0.000 | NaN | 2.000 | 2.000 | 0.500 | NaN | 0.000 | NaN | 2.000 | 2.000 |

| | Direct | | | | | | Indirect | | | | | | Reverse | | | | | |
|---|---|---|---|---|---|---|---|---|---|---|---|---|---|---|---|---|---|---|
| | $A$ | $P$ | $R$ | $F1$ | r-conf | nr-conf | $A$ | $P$ | $R$ | $F1$ | r-conf | nr-conf | $A$ | $P$ | $R$ | $F1$ | r-conf | nr-conf |
| gpt-3.5 | 0.305 | 0.206 | 0.137 | 0.165 | 8.203 | 7.492 | 0.317 | 0.388 | 0.634 | 0.482 | 8.602 | 8.923 | 0.499 | 0.499 | 0.721 | 0.590 | 7.797 | 7.764 |
| llama-3 | **0.620** | **0.572** | **0.947** | **0.714** | 8.567 | 8.038 | 0.484 | **0.491** | 0.887 | **0.632** | **10.000** | 9.985 | **0.605** | **0.955** | 0.221 | 0.359 | 8.496 | 8.473 |
| llama-2 | 0.497 | 0.485 | 0.084 | 0.143 | 8.000 | 8.000 | 0.138 | 0.216 | 0.276 | 0.243 | 8.000 | 8.000 | 0.503 | 0.501 | **0.995** | **0.667** | 8.000 | 8.000 |
| mistral | 0.500 | NaN | 0.000 | NaN | 9.000 | 9.000 | 0.347 | 0.392 | 0.553 | 0.459 | 9.000 | 9.000 | 0.500 | NaN | 0.000 | NaN | 7.534 | 7.797 |
| falcon | 0.395 | 0.367 | 0.289 | 0.324 | **10.000** | **10.000** | 0.462 | 0.480 | **0.924** | **0.632** | **10.000** | **10.000** | 0.514 | 0.649 | 0.063 | 0.115 | 9.783 | 9.724 |
| vicuna | 0.543 | 0.540 | 0.587 | 0.562 | 7.228 | 7.790 | 0.246 | 0.327 | 0.482 | 0.390 | 9.000 | 9.000 | 0.559 | 0.846 | 0.145 | 0.247 | 9.000 | 9.000 |
| neural-chat | 0.500 | NaN | 0.000 | NaN | **10.000** | **10.000** | 0.201 | 0.244 | 0.284 | 0.262 | **10.000** | **10.000** | 0.532 | 0.853 | 0.076 | 0.140 | **10.000** | **10.000** |
| gemma | 0.321 | 0.321 | 0.321 | 0.321 | 2.000 | 2.000 | **0.500** | NaN | 0.000 | NaN | 2.000 | 2.000 | 0.500 | NaN | 0.000 | NaN | 2.000 | 2.000 |

Table 5: Few-shot prompting results on HyperLex verb pairs: (top) HyperLex-FS, (bottom) Fellbaum-FS.

### 5.2 Evaluation on HyperLex data

**Zero-Shot Prompting.** Looking at the results in Table 4, we notice a certain consistency with the zero-shot results on WordNet verb pairs (Table 2) in terms of best-performing models for each of the assessment criteria: Gemma and NeuralChat under the Direct prompting, Mistral and Llama-2 under the Indirect prompting, Vicuna and NeuralChat but also GPT-3.5 under the Reverse prompting. Considering all models and promptings, results are substantially comparable to those achieved on WordNet data; however, in several cases, the models' performances are lower than on WordNet data, which would be explained by the lack of verb definitions in the prompts that use HyperLex verb pairs.

**Prompting with HyperLex-FS.** The HyperLex-FS based results (Table 5-top) offer a view which again resembles analogous results on WordNet data. In fact, relative improvements w.r.t. the zero-shot scenario occur for Llama-3, GPT-3.5 and Falcon under Direct and Indirect, and Vicuna under Direct, whereas most models cannot take advantage of the examples under Reverse.

**Prompting with Fellbaum-FS.** While no significant differences are observed in terms of the models' confidence, the Fellbaum-FS based results (Table 5-bottom) allow us to draw remarks that are similar in some cases to the corresponding evaluation on WordNet data. Particularly, compared to HyperLex-FS based results, Vicuna and, to a less extent, GPT-3.5 performance increases under Direct and Indirect, whereas Llama-3 and Falcon benefit from Fellbaum-FS only under the Reverse and Indirect prompting, respectively. Mistral also takes advantage of the Fellbaum-FS strategy under the Indirect prompting. Overall, Llama-3 remains the best performing model on the few-shot scenarios.

**Summary.** Our evaluation on HyperLex data has shown behaviors of the models that are close to those observed on WordNet data. Apart from the generally lower values compared to the corresponding zero- and few-shot scenarios on WordNet data, there is a certain consistency of the models in terms of their more favorable or unfavorable settings according to the performance criteria as well as in terms of their self-confidence.

## 6 Conclusions

We presented an investigation of the abilities of a representative body of LLMs in tackling the task of lexical entailment recognition. To the best of our knowledge, this is the first systematic study that aims to shed light on the actual skills of LLMs, with emphasis on Open models, to recognize entailment relations between verbs, gauging their accuracy against well-grounded, manually curated lexical resources such as WordNet and HyperLex. To accomplish our research goal, we defined three prompt types, providing different levels of contextual information, in both zero-shot and few-shot learning scenarios. Our results have shown evidence of both abilities and limitations that arise in the examined LLMs, which can be summarized as follows: $(i)$ although at varying degree of effectiveness and under different conditions, the LLMs can tackle a task of lexical entailment recognition with moderately good results, however, perfectly solving the task remains an unmet challenge for all the examined LLMs; $(ii)$ few-shot prompting can improve the models' performance in addressing the task; and $(iii)$ providing models with examples of entailment relation based on the Fellbaum types represents the best few-shot prompting strategy. Limitations and ethical considerations are also discussed next.

We believe our study can advance our understanding of how LLMs grasp nuanced meanings and logical relationships among verbs, providing valuable insights into their interpretability and decision-making processes.

## Acknowledgements

AT, resp. LLC, was supported by project "Future Artificial Intelligence Research (FAIR)" spoke 9 (H23C22000860006), resp. project SERICS (PE00000014), both under the MUR National Recovery and Resilience Plan funded by the EU - NextGenerationEU.

## Limitations

**Model types.** This work is focused on *general-purpose* models, as to date they represent the most widely used family of models in various NLP tasks. Nonetheless, models specifically designed for natural language understanding or inference might provide important insights into how LLMs address lexical entailment recognition. Future work might extend the scope of our evaluation to such models. Nonetheless, we would like to point out that at a late stage of writing of this paper we came across a recent study which regards a fine-tuned Llama-2 model for multiple lexical semantic tasks (Moskvoretskii et al., 2024). In Appendix E, we provide preliminary results concerning this model.

**Data resources.** Our findings are based on WordNet and HyperLex data on verb relations. Although they are invaluable lexical resources, they cannot be regarded as exhaustive in capturing all nuances of verb entailment relations; for example, they miss the specificity of sublanguages associated with particular domains, such as those of scientific fields or the legal domain. It would be interesting to extend our study to lexical entailment relations that characterize specialized domains as well.

**Broader Entailment Scope.** By referring to a broader context than lexical entailment, it is desirable to extend our study to embrace textual entailment as well. This might not be straightforward however, since, by requiring assessing the logical relationship between entire sentences or texts, textual entailment may involve multiple lexical entailment relationships within sentences.

**Restrictions on Closed LLMs.** The *Guidance* framework requires full access to the models to enforce grammar constraints effectively, such as with the `select` method. This works well with Open LLM, while closed LLMs, like the examined GPT-3.5, are only accessible via remote APIs, and hence do not support full integration with *Guidance*. Therefore, we avoided using a fixed grammar based on the `select` method and allowed GPT-3.5 to generate answers while enforcing it to meet our required format. To ensure valid outcomes by GPT-3.5, we eventually parsed the generated answers to extract the actual responses.

**Language usage.** Our evaluation refers to English verbs only. Results may differ in other languages, and extending the test to multiple languages using multilingual capable models could reveal variations in outcomes based on linguistic differences.

## Ethics Statement

**Broader impact.** The primary objective of our research is to advance the comprehension of how LLMs approach the task of lexical entailment, while also investigating how various prompting techniques harness the capabilities of these models. Our results indicate that certain models demonstrate robust NLU and NLI abilities. Although we believe our findings could facilitate a more profound and effective integration of LLMs into similar tasks, we decline any responsibility for any potential misuse or malicious applications stemming from our findings. Additionally, we emphasize the importance of all stakeholders exercising caution and responsibility to guarantee the safe and ethical implementation and utilization of these remarkable skills.

**Fair treatment of the models.** We ensured fairness in how the LLMs were evaluated, since all of them were given exactly the same prompts: nonetheless, one should recall that, by construction, each LLM features its own instruction template, and this template has to necessarily be followed as an input format when prompting the model in order to get response from it. In other terms, LLMs might require different input formats for handling a conversation, which is converted into a tokenizable string in the format that each model expects. Accordingly, we strictly adhere to each LLM's usage instructions, therefore our evaluation was carried out not disadvantaging any model.

## A Zero-shot Prompts on WordNet data

**System Prompt**

*You are a linguistic expert who responds to user questions in a concise way.*

**Direct Prompt**

*Given the verb $v_1$ defined as $def(v_1)$ and the verb $v_2$ defined as $def(v_2)$, what is the verb that entails the other? Answer must be either of the form "X entails Y" or "there is no entailment".*

*Answer:* `select{"`$v_1$` entails `$v_2$`", "`$v_2$` entails `$v_1$`", "there is not entailment"}.`

*Confidence of the answer (1 is the lowest, 10 is the highest):* `select{"1: No Confidence", "2: Very Low Confidence", "3: Low Confidence", "4: Fair Confidence", "5: Moderate Confidence", "6: Good Confidence", "7: High Confidence", "8: Very High Confidence", "9: Extremely High Confidence", "10: Absolute Certainty"}.`

**Indirect Prompt**

*Relation F states that given two verbs X and Y, X and Y satisfy F if and only if when doing Y you are also doing X.*

*Given the verb $v_1$ defined as $def(v_1)$ and the verb $v_2$ defined as $def(v_2)$, what is X and what is Y? Answer must be either a pair (X,Y) or "relation F cannot be satisfied".*

*Answer:* `select{"(`$v_1$`, `$v_2$`)", "(`$v_2$`, `$v_1$`)", "relation F cannot be satisfied"}.`

*Confidence of the answer (1 is the lowest, 10 is the highest):* `select{"1: No Confidence", "2: Very Low Confidence", "3: Low Confidence", "4: Fair Confidence", "5: Moderate Confidence", "6: Good Confidence", "7: High Confidence", "8: Very High Confidence", "9: Extremely High Confidence", "10: Absolute Certainty"}.`

**Reverse Prompt**

*Given the verb $v_1$ defined as $def(v_1)$, and the verb $v_2$ defined as $def(v_2)$, answer YES if 'not $v_2$' entails 'not $v_1$,' NO otherwise.*

*Answer:* `select{"YES", "NO"}.`

*Confidence of the answer (1 is the lowest, 10 is the highest):* `select{"1: No Confidence", "2: Very Low Confidence", "3: Low Confidence", "4: Fair Confidence", "5: Moderate Confidence", "6: Good Confidence", "7: High Confidence", "8: Very High Confidence", "9: Extremely High Confidence", "10: Absolute Certainty"}.`

## B Zero-shot Prompts on HyperLex data

**System Message**

*You are a linguistic expert who responds to user questions in a concise way.*

**Direct Prompt**

*Given the verb $v_1$ and the verb $v_2$, what is the verb that entails the other? Answer must be either of the form "X entails Y" or "there is no entailment".*

*Answer:* `select{"`$v_1$` entails `$v_2$`", "`$v_2$` entails `$v_1$`", "there is not entailment"}.`

*Confidence of the answer (1 is the lowest, 10 is the highest):* `select{"1: No Confidence", "2: Very Low Confidence", "3: Low Confidence", "4: Fair Confidence", "5: Moderate Confidence", "6: Good Confidence", "7: High Confidence", "8: Very High Confidence", "9: Extremely High Confidence", "10: Absolute Certainty"}.`

**Indirect Prompt**

*Relation F states that given two verbs X and Y, X and Y satisfy F if and only if when doing Y you are also doing X.*

*Given the verb $v_1$ and the verb $v_2$, what is X and what is Y? Answer must be either a pair (X,Y) or "relation F cannot be satisfied".*

*Answer:* `select{"(`$v_1$`, `$v_2$`)", "(`$v_2$`, `$v_1$`)", "relation F cannot be satisfied"}.`

*Confidence of the answer (1 is the lowest, 10 is the highest):* `select{"1: No Confidence", "2: Very Low Confidence", "3: Low Confidence", "4: Fair Confidence", "5: Moderate Confidence", "6: Good Confidence", "7: High Confidence", "8: Very High Confidence", "9: Extremely High Confidence", "10: Absolute Certainty"}.`

**Reverse Prompt**

*Given the verb $v_1$ and the verb $v_2$, answer YES if 'not $v_2$' entails 'not $v_1$,' NO otherwise.*

*Answer:* `select{"YES", "NO"}`.

*Confidence of the answer (1 is the lowest, 10 is the highest):* `select{"1: No Confidence", "2: Very Low Confidence", "3: Low Confidence", "4: Fair Confidence", "5: Moderate Confidence", "6: Good Confidence", "7: High Confidence", "8: Very High Confidence", "9: Extremely High Confidence", "10: Absolute Certainty"}`.

## C Few-shot prompt selection strategy

For the HyperLex-FS prompting, we selected the following verb pairs, one from each of the subranges within [2..6] with step 1 (the associated HyperLex score is also reported within square brackets):

- (warn, advise) [5.75]
- (instruct, inform) [4.31]
- (rationalize, argue) [3.08]
- (take, have) [2.17]

Analogously, for the Fellbaum-FS prompting, we selected the following verb pairs, one from each of the Fellbaum's categories of verb entailments (Fellbaum, 1990):

- *Entailment with temporal co-extensiveness*: the activity denoted by the entailing verb implies a more general one, denoted by the entailed verb, in a simultaneous manner (i.e., they are temporally co-extensive). This corresponds to troponymy.
- *Entailment with temporal proper inclusion*: the activity denoted by the entailing verb includes the activity denoted by the entailed verb, and is not temporally co-extensive (i.e., one activity can occur before or after the other).
- *Entailment with backward presupposition*: the activity denoted by the entailed verb always precedes the activity denoted by the entailing verb in time.
- *Cause*: if the activity denoted by verb $v_1$ causes the activity denoted by verb $v_2$, then $v_1$ also entails $v_2$.

It should be noted that no categorization of the WordNet verbs according to the Fellbaum taxonomy was carried out, since WordNet does not provide annotations on the Fellbaum taxonomy types. In addition, evaluating a prediction task on the Fellbaum taxonomy types would be challenging since we are not aware of the existence of lexical databases that provide ground truth labels for these subtypes, thus going beyond the scope of this study.

Below we show the Fellbaum-FS-based verb pairs that were selected to define four response examples to provide to each of the models in a few-shot scenario:

- (limp, walk), as example of troponymy co-extensiveness relation;
- (snore, sleep), as example of troponymy proper inclusion relation;
- (succeed, try), as example of backward presupposition relation;
- (give, have), as example of cause relation.

In the following, we report the pre-prompts (user messages) used for each of the three prompt types (i.e., Direct, Indirect, and Reverse) with HyperLex-FS examples and Fellbaum-FS examples, respectively, over WordNet data. Note that these pre-prompts also apply to HyperLex data apart from the specification of verb definitions.

### HyperLex-FS pre-prompt for the Direct prompt

Here are some examples of the task you have to perform:

*Start Examples*

- Example 1
Input: Given the verb 'warn' defined as 'admonish or counsel in terms of someone's behavior', and the verb 'advise' defined as 'give advice to'. What is the verb that entails the other? Answer must be of the form X entails Y.

Output: 'warn' entails 'advise'

- Example 2
Input: Given the verb 'inform' defined as 'impart knowledge of some fact, state or affairs, or event to', and the verb 'instruct' defined as 'make aware of'. What is the verb that entails the other? Answer must be of the form X entails Y.

Output: 'instruct' entails 'inform'

- Example 3
Input: Given the verb 'argue' defined as 'present reasons and arguments', and the verb 'rationalize' defined as 'defend, explain, clear away, or make excuses for by reasoning'. What is the verb that entails the other? Answer must be of the form X entails Y.

Output: 'rationalize' entails 'argue'

- Example 4
Input: Given the verb 'take' defined as 'experience or feel or submit to', and the verb 'have' defined as 'go through (mental or physical states or experiences)'. What is the verb that entails the other? Answer must be of the form X entails Y.

Output: 'take' entails 'have'

*End Examples*

### HyperLex-FS pre-prompt for the Indirect prompt

Here are some examples of the task you have to perform:

*Start Examples*

- Example 1
Input: Relation F states that given two verbs X and Y, X and Y satisfy F if and only if when doing Y you are also doing X. Given the verb 'warn' defined as 'admonish or counsel in terms of someone's behavior', and the verb 'advise' defined as 'give advice to' What is X and what is Y? Answer must be a pair (X ,Y).

Output: (warn, advise)

- Example 2
Input: Relation F states that given two verbs X and Y, X and Y satisfy F if and only if when doing Y you are also doing X. Given the verb 'inform' defined as 'impart knowledge of some fact, state or affairs, or event to', and the verb 'instruct' defined as 'make aware of' What is X and what is Y? Answer must be a pair (X ,Y).

Output: (instruct, inform)

- Example 3
Input: Relation F states that given two verbs X and Y, X and Y satisfy F if and only if when doing Y you are also doing X. Given the verb 'argue' defined as 'present reasons and arguments', and the verb 'rationalize' defined as 'defend, explain, clear away, or make excuses for by reasoning' What is X and what is Y? Answer must be a pair (X ,Y).

Output: (rationalize, argue)

- Example 4
Input: Relation F states that given two verbs X and Y, X and Y satisfy F if and only if when doing Y you are also doing X. Given the verb 'take' defined as 'experience or feel or submit to', and the verb 'have' defined as 'go through (mental or physical states or experiences)' What is X and what is Y? Answer must be a pair (X ,Y).

Output: (take, have)

*End Examples*

## HyperLex-FS pre-prompt for the Reverse prompt

Here are some examples of the task you have to perform:

*Start Examples*

- Example 1

Input: Given the verb 'warn' defined as 'admonish or counsel in terms of someone's behavior', and the verb 'advise' defined as 'give advice to' Answer YES if 'not warn' entails 'not advise', NO otherwise.

Output: NO

- Example 2

Input: Given the verb 'inform' defined as 'impart knowledge of some fact, state or affairs, or event to', and the verb 'instruct' defined as 'make aware of' Answer YES if 'not inform' entails 'not instruct', NO otherwise.

Output: YES

- Example 3

Input: Given the verb 'argue' defined as 'present reasons and arguments', and the verb 'rationalize' defined as 'defend, explain, clear away, or make excuses for by reasoning' Answer YES if 'not argue' entails 'not rationalize', NO otherwise.

Output: YES

- Example 4

Input: Given the verb 'take' defined as 'experience or feel or submit to', and the verb 'have' defined as 'go through (mental or physical states or experiences)' Answer YES if 'not take' entails 'not have', NO otherwise.

Output: NO

*End Examples*

## Fellbaum-FS pre-prompt for the Direct prompt

Here are some examples of the task you have to perform:

*Start Examples*

- Example 1

Input: Given the verb 'limp' defined as 'walk impeded by some physical limitation or injury', and the verb 'walk' defined as 'use one's feet to advance; advance by steps'. What is the verb that entails the other? Answer must be of the form X entails Y.

Output: 'limp' entails 'walk'

- Example 2

Input: Given the verb 'sleep' defined as 'be asleep', and the verb 'snore' defined as 'breathe noisily during one's sleep'. What is the verb that entails the other? Answer must be of the form X entails Y.

Output: 'snore' entails 'sleep'

- Example 3

Input: Given the verb 'try' defined as 'make an effort or attempt', and the verb 'succeed' defined as 'attain success or reach a desired goal'. What is the verb that entails the other? Answer must be of the form X entails Y.

Output: 'succeed' entails 'try'

- Example 4

Input: Given the verb 'give' defined as 'guide or direct, as by behavior of persuasion', and the verb 'have' defined as 'cause to do; cause to act in a specified manner'. What is the verb that entails the other? Answer must be of the form X entails Y.

Output: 'give' entails 'have'

*End Examples*

**Fellbaum-FS pre-prompt for the Indirect prompt**

Here are some examples of the task you have to perform:

*Start Examples*

- Example 1

Input: Relation F states that given two verbs X and Y, X and Y satisfy F if and only if when doing Y you are also doing X. Given the verb 'limp' defined as 'walk impeded by some physical limitation or injury', and the verb 'walk' defined as 'use one's feet to advance; advance by steps' What is X and what is Y? Answer must be a pair (X ,Y).

Output: (limp, walk)

- Example 2

Input: Relation F states that given two verbs X and Y, X and Y satisfy F if and only if when doing Y you are also doing X. Given the verb 'sleep' defined as 'be asleep', and the verb 'snore' defined as 'breathe noisily during one's sleep' What is X and what is Y? Answer must be a pair (X ,Y).

Output: (sleep, snore)

- Example 3

Input: Relation F states that given two verbs X and Y, X and Y satisfy F if and only if when doing Y you are also doing X. Given the verb 'try' defined as 'make an effort or attempt', and the verb 'succeed' defined as 'attain success or reach a desired goal' What is X and what is Y? Answer must be a pair (X ,Y).

Output: (succeed, try)

- Example 4

Input: Relation F states that given two verbs X and Y, X and Y satisfy F if and only if when doing Y you are also doing X. Given the verb 'give' defined as 'guide or direct, as by behavior of persuasion', and the verb 'have' defined as 'cause to do; cause to act in a specified manner' What is X and what is Y? Answer must be a pair (X ,Y).

Output: (give, have)

*End Examples*

**Fellbaum-FS pre-prompt for the Reverse prompt**

Here are some examples of the task you have to perform:

*Start Examples*

- Example 1

Input: Given the verb 'limp' defined as 'walk impeded by some physical limitation or injury', and the verb 'walk' defined as 'use one's feet to advance; advance by steps' Answer YES if 'not limp' entails 'not walk', NO otherwise.

Output: NO

- Example 2

Input: Given the verb 'sleep' defined as 'be asleep', and the verb 'snore' defined as 'breathe noisily during one's sleep'. Answer YES if 'not sleep' entails 'not snore', NO otherwise.

Output: YES

- Example 3

Input: Given the verb 'try' defined as 'make an effort or attempt', and the verb 'succeed' defined as 'attain success or reach a desired goal'. Answer YES if 'not try' entails 'not succeed', NO otherwise.

Output: YES

- Example 4

Input: Given the verb 'give' defined as 'guide or direct, as by behavior of persuasion', and the verb 'have' defined as 'cause to do; cause to act in a specified manner'. Answer YES if 'not give' entails 'not have', NO otherwise.

Output: NO

*End Examples*

# D Details on the assessment criteria

Table 6 reports the definitions of true positives (TP), true negatives (TN), false positives (FP), and false negatives (FN) for each of our defined types of prompt. Symbols $p^{\mathsf{rel}}(X, Y)$and $p^{\neg\mathsf{rel}}(X, Y)$are used to an input relevant and not-relevant verb-pair, respectively.

| | Direct prompt type | | Indirect prompt type | | Reverse prompt type | |
|---|---|---|---|---|---|---|
| | input | answer | input | answer | input | answer |
| TP | $p^{\text{rel}}(X,Y)$ | X entails Y | $p^{\text{rel}}(X,Y)$ | $(X,Y)$ | $p^{\text{rel}}(X,Y)$ | Yes |
| TN | $p^{\neg\text{rel}}(X,Y)$ | there is no entailment | $p^{\neg\text{rel}}(X,Y)$ | relation F cannot be satisfied | $p^{\neg\text{rel}}(X,Y)$ | No |
| FP | $p^{\neg\text{rel}}(X,Y)$ | $X$ entails $Y$, or $Y$ entails $X$ | $p^{\neg\text{rel}}(X,Y)$ | $(X,Y)$ or $(Y,X)$ | $p^{\neg\text{rel}}(X,Y)$ | Yes |
| FN | $p^{\text{rel}}(X,Y)$<br>$p^{\text{rel}}(X,Y)$ | there is no entailment<br>$Y$ entails $X$ | $p^{\text{rel}}(X,Y)$<br>$p^{\text{rel}}(X,Y)$ | relation F cannot be satisfied<br>$(Y,X)$ | $p^{\text{rel}}(X,Y)$ | No |

Table 6: Description of the confusion matrix statistics for each of the three prompt types.

# E Further experimental results

**Impact of WordNet relation types.** As we discussed in Sect. 4.1, WordNet verb pairs involved in entailment relations can be accessed through two methods, namely *hyponyms()* and *entailments()*.

In Table 7, we summarize results on our evaluation of the impact of the two methods on the LLMs' responses, where the values reported in the table correspond to percentage values of the correctly identified verb pairs. Results show no significant differences in the percentage of correctly recognized verb relations for the best-performing models, for each prompt type and zero-/few-shot scenario. On average over all models, the set of correctly recognized verb relations of both types tends to be larger in the zero-shot scenario (around 33% for Direct, 68% for Indirect, and 60% for Reverse) followed by Fellbaum-FS (around 32% for Direct, 56% for Indirect, and 39% for Reverse).

| | Zero-shot | | | | | | HyperLex-FS | | | | | | Fellbaum-FS | | | | | |
|---|---|---|---|---|---|---|---|---|---|---|---|---|---|---|---|---|---|---|
| | Direct | | Indirect | | Reverse | | Direct | | Indirect | | Reverse | | Direct | | Indirect | | Reverse | |
| | *Ent.* | *Hypo.* | *Ent.* | *Hypo.* | *Ent.* | *Hypo.* | *Ent.* | *Hypo.* | *Ent.* | *Hypo.* | *Ent.* | *Hypo.* | *Ent.* | *Hypo.* | *Ent.* | *Hypo.* | *Ent.* | *Hypo.* |
| `gpt-3.5` | 18.35 | 12.45 | 17.40 | 15.30 | 96.65 | 96.25 | 18.05 | 15.65 | 78.25 | 80.90 | **90.30** | **87.75** | 19.15 | 18.20 | 82.05 | 84.30 | 81.85 | 76.05 |
| `llama-3` | 24.55 | 15.35 | 56.45 | 61.40 | 75.55 | 76.40 | 42.40 | 38.50 | 73.95 | 72.60 | 26.30 | 24.95 | **85.25** | **86.55** | 74.85 | 75.65 | 31.10 | 25.65 |
| `llama-2` | 21.65 | 20.60 | 52.00 | 49.15 | 19.40 | 24.80 | 14.00 | 14.45 | 9.65 | 7.60 | 31.20 | 25.30 | 23.75 | 24.35 | 24.70 | 22.95 | 69.65 | 58.05 |
| `mistral` | 22.55 | 20.75 | **92.40** | **93.60** | 0.00 | 0.00 | 0.00 | 0.00 | 33.55 | 36.25 | 0.00 | 0.00 | 0.00 | 0.00 | 27.25 | 24.20 | 0.00 | 0.00 |
| `falcon` | 17.50 | 16.70 | 82.15 | 82.90 | 85.75 | 91.45 | 38.50 | 39.85 | **87.05** | **89.40** | 52.00 | 50.70 | 38.30 | 39.45 | **90.35** | **92.15** | **94.35** | **94.35** |
| `vicuna` | 49.55 | 51.70 | 89.45 | 88.55 | **100.00** | **100.00** | **54.65** | **55.90** | 72.50 | 70.20 | 36.55 | 34.00 | 56.20 | 58.75 | 65.75 | 67.15 | 35.00 | 29.00 |
| `neural-chat` | 28.45 | 26.45 | 66.20 | 73.65 | 29.10 | 31.15 | 0.00 | 0.00 | 64.90 | 59.35 | 15.05 | 17.65 | 0.00 | 0.00 | 83.25 | 80.65 | 18.10 | 16.35 |
| `gemma` | **86.90** | **89.40** | 80.50 | 83.50 | 62.15 | 63.70 | 31.65 | 31.50 | 0.00 | 0.00 | 0.00 | 0.00 | 31.65 | 31.50 | 0.00 | 0.00 | 0.00 | 0.00 |

Table 7: Percentage of WordNet *hyponyms()*, resp. *entailments()*, based verb-pairs correctly recognized in entailment relations by the models, for each prompt type and strategy

**Distribution of the lexicographers' files over all verbs involved in entailment relations.** WordNet verbs can also be categorized based on the 15 lexicographers' files for verbs. In the NLTK WordNet library, these verb categories can be accessed through the *lexname()* function defined over synsets.

We carried out an additional statistical analysis focused on the distribution of the 15 lexicographers' files over all verbs involved in entailment relations. We present our developed methodology as follows.

For each model M, for each prompt type P (i.e., Direct, Indirect, Reverse), and for each relation in entailments(), hyponyms(), in both zero-shot and few-shot settings, we calculated:

- the distribution of *lexname* categories associated to all entailing verbs X, for which the model M correctly answered the prompt P on the entailment relation (X,Y).
- the distribution of *lexname* categories associated to all entailed verbs Y, for which the model M correctly answered the prompt P on the entailment relation (X,Y).
- the distribution of *lexname* categories involved in those verb pairs for which the model M correctly answered the prompt P on the entailment relation (X,Y), such that the *lexname* of X coincides with the *lexname* of Y.

For each of these distributions, we calculated the distribution entropy and inspected the most frequently occurring *lexname* categories.

The above analysis was analogously repeated for all verb pairs involving entailment relations that were not correctly recognized by a model. In Table 8, we present a summary of the results obtained on the zero-shot scenario — analogous results were observed on the two few-shot scenarios as well. Specifically, we report the following information: under the "top-3 lexnames" column, we report for each model and setting, the three *lexname* categories that are most frequently associated with either verb in a pair, over all prompting types, whereas the "avg. norm. entropy" column refers to the average (over all prompting types) of the normalized entropy values of the distributions of *lexname* categories. We also note that (results not shown) the 15 *lexname* categories are always represented in each of the distributions, regardless of the response type, which indicates that no subset of verb categories characterizes either the 'correct' or the 'wrong' answers.

| **model** | **relation** | **outcome** | **top-3 lexnames** | **avg. norm. entropy** |
|---|---|---|---|---|
| gpt-3.5 | entailments() | correct | contact, motion, communication | 0.89 |
| gpt-3.5 | entailments() | wrong | contact, motion, communication | 0.91 |
| gpt-3.5 | hyponyms() | correct | change, contact, communication | 0.90 |
| gpt-3.5 | hyponyms() | wrong | change, communication, contact | 0.91 |
| llama-3 | entailments() | correct | contact, motion, communication | 0.91 |
| llama-3 | entailments() | wrong | contact, change, communication | 0.93 |
| llama-3 | hyponyms() | correct | change, contact, communication | 0.91 |
| llama-3 | hyponyms() | wrong | change, contact, communication | 0.91 |
| llama-2 | entailments() | correct | contact, motion, communication | 0.91 |
| llama-2 | entailments() | wrong | contact, motion, communication | 0.91 |
| llama-2 | hyponyms() | correct | contact, change, communication | 0.89 |
| llama-2 | hyponyms() | wrong | contact, change, communication | 0.90 |
| mistral | entailments() | correct | contact, motion, communication | 0.91 |
| mistral | entailments() | wrong | contact, body, motion | 0.91 |
| mistral | hyponyms() | correct | change, communication, stative | 0.90 |
| mistral | hyponyms() | wrong | communication, contact, motion | 0.93 |
| falcon | entailments() | correct | contact, motion, communication | 0.92 |
| falcon | entailments() | wrong | contact, motion, communication | 0.92 |
| falcon | hyponyms() | correct | contact, change, communication | 0.92 |
| falcon | hyponyms() | wrong | contact, change, communication | 0.90 |
| vicuna | entailments() | correct | contact, motion, communication | 0.94 |
| vicuna | entailments() | wrong | motion, communication, contact | 0.91 |
| vicuna | hyponyms() | correct | change, communication, contact | 0.89 |
| vicuna | hyponyms() | wrong | change, communication, contact | 0.90 |
| neural-chat | entailments() | correct | communication, motion, contact | 0.91 |
| neural-chat | entailments() | wrong | communication, motion, contact | 0.92 |
| neural-chat | hyponyms() | correct | change, motion, contact | 0.89 |
| neural-chat | hyponyms() | wrong | communication, contact, change | 0.88 |
| gemma | entailments() | correct | motion, communication, contact | 0.92 |
| gemma | entailments() | wrong | change, contact, communication | 0.91 |
| gemma | hyponyms() | correct | contact, motion, communication | 0.91 |
| gemma | hyponyms() | wrong | change, contact, communication | 0.93 |

Table 8: Distribution of the verb lexicographers' files: zero-shot scenario

Looking at the table at first glance, there is evidence of a small subset of lexname categories that consistently appear in the top-3 column; this is actually not surprising since those correspond to the most represented lexname categories among the verbs in WordNet, namely 'change', 'contact', 'communication', 'motion' and 'social', with more than 60% of the verb synsets falling into one of these categories. More importantly, by comparing the results corresponding to 'correct' and 'wrong' responses, the top-3 categories are mostly overlapping; this holds consistently for each model, regardless of the relation type and whether few-shots were used in the prompts. Also, for each model and relation type, the values of average normalized entropy (ranging within [0,1]) of the lexname distributions of 'correct' and 'wrong' responses are equally very high and similar to each other, thus hinting at a common trait of heterogeneity of lexname categories occurring in verb pairs corresponding to valid (either correct or wrong) responses.

Overall, based on this empirical evidence, we can conclude that the lexname categories cannot be regarded as predictors of a model's performance in recognizing verb entailments; therefore, there is no evidence that LLMs fail to understand a particular subset of verbs in recognizing lexical entailments.

**Preliminary experiments on TaxoLLaMA.** TaxoLLaMA has been very recently (in March 2024)

| | Direct | | | | | | Indirect | | | | | | Reverse | | | | | |
|---|---|---|---|---|---|---|---|---|---|---|---|---|---|---|---|---|---|---|
| | $A$ | $P$ | $R$ | $F1$ | r-conf | nr-conf | $A$ | $P$ | $R$ | $F1$ | r-conf | nr-conf | $A$ | $P$ | $R$ | $F1$ | r-conf | nr-conf |
| llama-3 | **0.571** | **0.862** | 0.169 | **0.283** | **9.94** | **9.95** | 0.334 | 0.392 | 0.605 | 0.476 | **9.07** | **8.65** | **0.605** | 0.579 | 0.767 | 0.660 | **9.74** | **9.67** |
| llama-2 | 0.422 | 0.363 | 0.205 | 0.262 | 8.00 | 8.00 | **0.670** | **0.623** | 0.860 | **0.723** | 7.98 | 8.00 | 0.539 | **0.599** | 0.236 | 0.339 | 7.99 | 7.99 |
| taxollama | 0.167 | 0.248 | **0.329** | **0.283** | 9.00 | 9.00 | *0.500* | *NaN* | *0.000* | *NaN* | *10.00* | *10.00* | 0.500 | 0.500 | **1.000** | **0.667** | 9.00 | 9.00 |

Table 9: Comparison of TaxoLlama with Llama-2 and Llama-3 (cf. Table 2): Zero-shot prompting results on WordNet verb pairs.

| | Direct | | | | | | Indirect | | | | | | Reverse | | | | | |
|---|---|---|---|---|---|---|---|---|---|---|---|---|---|---|---|---|---|---|
| | $A$ | $P$ | $R$ | $F1$ | r-conf | nr-conf | $A$ | $P$ | $R$ | $F1$ | r-conf | nr-conf | $A$ | $P$ | $R$ | $F1$ | r-conf | nr-conf |
| llama-3 | **0.660** | **0.831** | 0.402 | **0.542** | **9.71** | **9.30** | **0.634** | **0.606** | **0.767** | **0.677** | **9.77** | **9.27** | **0.617** | **0.958** | **0.244** | **0.389** | 8.94 | 8.71 |
| llama-2 | 0.358 | 0.256 | 0.149 | 0.188 | 8.01 | 8.00 | 0.048 | 0.087 | 0.095 | 0.091 | 8.00 | 8.00 | 0.539 | 0.602 | 0.231 | 0.333 | 8.00 | 8.00 |
| taxollama | 0.284 | 0.343 | **0.474** | 0.398 | 9.10 | 9.04 | 0.265 | 0.305 | 0.367 | 0.333 | 9.31 | 9.09 | 0.514 | 0.567 | 0.121 | 0.200 | 9.01 | 9.00 |
| | **Direct** | | | | | | **Indirect** | | | | | | **Reverse** | | | | | |
| | $A$ | $P$ | $R$ | $F1$ | r-conf | nr-conf | $A$ | $P$ | $R$ | $F1$ | r-conf | nr-conf | $A$ | $P$ | $R$ | $F1$ | r-conf | nr-conf |
| llama-3 | **0.661** | **0.617** | **0.852** | **0.715** | **9.29** | 8.54 | **0.508** | **0.505** | **0.761** | **0.607** | **9.87** | **9.57** | **0.617** | **0.942** | 0.251 | 0.396 | **9.23** | **9.03** |
| llama-2 | 0.290 | 0.271 | 0.248 | 0.259 | 8.00 | 8.00 | 0.120 | 0.194 | 0.240 | 0.215 | 8.00 | 8.00 | 0.526 | 0.523 | **0.585** | **0.552** | 8.00 | 8.00 |
| taxollama | 0.328 | 0.366 | 0.469 | 0.411 | 9.26 | **9.10** | 0.276 | 0.343 | 0.489 | 0.403 | 9.11 | 9.00 | 0.528 | 0.586 | 0.194 | 0.291 | 9.01 | 9.01 |

Table 10: Comparison of TaxoLlama with Llama-2 and Llama-3 (cf. Table 3): Few-shot prompting results on WordNet verb pairs: (top) HyperLex-FS, (bottom) Fellbaum-FS.

| | Direct | | | | | | Indirect | | | | | | Reverse | | | | | |
|---|---|---|---|---|---|---|---|---|---|---|---|---|---|---|---|---|---|---|
| | $A$ | $P$ | $R$ | $F1$ | r-conf | nr-conf | $A$ | $P$ | $R$ | $F1$ | r-conf | nr-conf | $A$ | $P$ | $R$ | $F1$ | r-conf | nr-conf |
| llama-3 | **0.405** | **0.409** | **0.426** | **0.418** | **9.075** | 8.476 | 0.271 | 0.352 | **0.542** | **0.427** | 9.325 | 8.996 | **0.625** | 0.673 | 0.487 | 0.565 | **9.648** | **9.425** |
| llama-2 | 0.355 | 0.353 | 0.347 | 0.350 | 8.000 | 8.000 | **0.504** | **0.636** | 0.018 | 0.036 | 6.449 | 6.281 | 0.533 | **0.727** | 0.105 | 0.184 | 7.968 | 7.961 |
| taxollama | 0.259 | 0.341 | 0.518 | 0.412 | 9.000 | **9.000** | 0.499 | 0.000 | 0.000 | NaN | **10.000** | **9.997** | 0.500 | 0.500 | **1.000** | **0.667** | 9.003 | 9.000 |

Table 11: Comparison of TaxoLlama with Llama-2 and Llama-3 (cf. Table 4): Zero-shot prompting results on HyperLex verb pairs.

| | Direct | | | | | | Indirect | | | | | | Reverse | | | | | |
|---|---|---|---|---|---|---|---|---|---|---|---|---|---|---|---|---|---|---|
| | $A$ | $P$ | $R$ | $F1$ | r-conf | nr-conf | $A$ | $P$ | $R$ | $F1$ | r-conf | nr-conf | $A$ | $P$ | $R$ | $F1$ | r-conf | nr-conf |
| llama-3 | **0.711** | **0.666** | **0.845** | **0.745** | 8.896 | 8.286 | **0.557** | **0.537** | **0.826** | **0.651** | **10.000** | **9.947** | **0.551** | **0.915** | 0.113 | 0.201 | 8.048 | 8.035 |
| llama-2 | 0.504 | 0.516 | 0.126 | 0.203 | 8.000 | 8.000 | 0.042 | 0.078 | 0.084 | 0.081 | 8.000 | 8.000 | 0.499 | 0.499 | **0.997** | **0.665** | 8.000 | 8.000 |
| taxollama | 0.370 | 0.423 | 0.716 | 0.532 | **9.032** | **9.023** | 0.159 | 0.223 | 0.274 | 0.246 | 9.140 | 9.033 | 0.516 | 0.561 | 0.145 | 0.230 | **9.056** | **9.052** |
| | **Direct** | | | | | | **Indirect** | | | | | | **Reverse** | | | | | |
| | $A$ | $P$ | $R$ | $F1$ | r-conf | nr-conf | $A$ | $P$ | $R$ | $F1$ | r-conf | nr-conf | $A$ | $P$ | $R$ | $F1$ | r-conf | nr-conf |
| llama-3 | **0.620** | **0.572** | **0.947** | **0.714** | 8.567 | 8.038 | **0.484** | **0.491** | **0.887** | **0.632** | **10.000** | **9.985** | **0.605** | **0.955** | 0.221 | 0.359 | 8.496 | 8.473 |
| llama-2 | 0.497 | 0.485 | 0.084 | 0.143 | 8.000 | 8.000 | 0.138 | 0.216 | 0.276 | 0.243 | 8.000 | 8.000 | 0.503 | 0.501 | **0.995** | **0.667** | 8.000 | 8.000 |
| taxollama | 0.338 | 0.402 | 0.663 | 0.500 | **9.019** | **9.016** | 0.280 | 0.297 | 0.321 | 0.308 | 9.432 | 9.256 | 0.533 | 0.531 | 0.561 | 0.545 | **9.333** | **9.332** |

Table 12: Comparison of TaxoLlama with Llama-2 and Llama-3 (cf. Table 5): Few-shot prompting results on HyperLex verb pairs: (top) HyperLex-FS, (bottom) Fellbaum-FS.

proposed in (Moskvoretskii et al., 2024) as a lightweight fine-tune of LLaMA2-7b, which is designed to deal with multiple lexical semantics tasks with focus on taxonomy related tasks, including Taxonomy Enrichment, Hypernym Discovery, Taxonomy Construction, and Lexical Entailment tasks.

In Tables 9–12 we present the results of our evaluation of TaxoLLaMA on our WordNet and HyperLex datasets, with focus on its comparison with the Llama models previously included in our evaluation LLMs. Looking at the results from the tables, it stands out that TaxoLLaMA is not only outperformed by Llama-3 in nearly all cases, but also it might still behave worse than Llama-2 under certain conditions (e.g., zero-shot under Indirect on both datasets, HyperLex-FS under Reverse on WordNet data, and in other cases according to one or more assessment criteria). Interestingly, these findings contrast with the fact that TaxoLLaMA derives from a fine-tuning of Llama-2 on WordNet and related lexical taxonomies, and that in (Moskvoretskii et al., 2024) the model is said to show *"strong zero-shot performance on lexical entailment with no fine-tuning"*. Our results would hence indicate the need for a deeper investigation of TaxoLLaMA on verb lexical entailment tasks.

**Accuracy and Confidence w.r.t. the average score in HyperLex.** We investigated whether the accuracy and confidence of the models can vary w.r.t. how the HyperLex verb-pairs were associated with their scores. The HyperLex scores consist of the average of the scores assigned by human annotators, on a scale from 0 to 6, based on the degree to which a lexical entailment relationship holds for any two verbs.

| | $A$ | | | | | | r-conf | | | | | |
|---|---|---|---|---|---|---|---|---|---|---|---|---|
| | $bin0$ | $bin1$ | $bin2$ | $bin3$ | $bin4$ | $bin5$ | $bin0$ | $bin1$ | $bin2$ | $bin3$ | $bin4$ | $bin5$ |
| `gpt-3.5` | 0.333 | 0.344 | 0.363 | 0.352 | 0.384 | 0.393 | 8.667 | 8.290 | 8.607 | 8.167 | 8.630 | 8.501 |
| `llama-3` | 0.363 | 0.365 | 0.519 | 0.486 | 0.505 | 0.539 | 9.431 | 9.236 | 8.997 | 9.004 | 9.123 | 9.274 |
| `llama-2` | 0.196 | 0.188 | 0.163 | 0.129 | 0.153 | 0.155 | 8.000 | 8.000 | 8.000 | 8.000 | 8.000 | 8.000 |
| `mistral` | 0.461 | 0.396 | 0.378 | 0.386 | 0.437 | 0.438 | 9.000 | 9.000 | 9.000 | 9.000 | 9.000 | 9.000 |
| `falcon` | 0.549 | 0.521 | 0.548 | 0.548 | 0.497 | 0.525 | 10.000 | 10.000 | 10.000 | 10.000 | 10.000 | 10.000 |
| `vicuna` | 0.863 | 0.854 | 0.874 | 0.876 | 0.839 | 0.863 | 9.000 | 9.000 | 9.000 | 9.000 | 9.000 | 9.000 |
| `neural-chat` | 0.402 | 0.385 | 0.333 | 0.386 | 0.495 | 0.447 | 10.000 | 10.000 | 10.000 | 10.000 | 10.000 | 10.000 |
| `gemma` | 0.431 | 0.385 | 0.511 | 0.448 | 0.524 | 0.516 | 2.000 | 2.000 | 2.000 | 2.000 | 2.000 | 2.000 |
| `taxollama` | 0.500 | 0.490 | 0.511 | 0.529 | 0.492 | 0.516 | 9.000 | 9.016 | 9.000 | 9.000 | 9.000 | 9.000 |

Table 13: Zero-shot prompting based accuracy and confidence over correct answers w.r.t. the HyperLex score bins.

| | $A$ | | | | | | r-conf | | | | | |
|---|---|---|---|---|---|---|---|---|---|---|---|---|
| | $bin0$ | $bin1$ | $bin2$ | $bin3$ | $bin4$ | $bin5$ | $bin0$ | $bin1$ | $bin2$ | $bin3$ | $bin4$ | $bin5$ |
| `gpt-3.5` | 0.324 | 0.552 | 0.548 | 0.519 | 0.595 | 0.607 | 8.568 | 8.014 | 7.987 | 7.781 | 7.910 | 7.970 |
| `llama-3` | 0.500 | 0.469 | 0.511 | 0.600 | 0.619 | 0.699 | 9.321 | 9.182 | 9.402 | 8.882 | 8.844 | 8.887 |
| `llama-2` | 0.441 | 0.375 | 0.415 | 0.386 | 0.389 | 0.429 | 8.000 | 8.000 | 8.000 | 8.000 | 8.000 | 8.000 |
| `mistral` | 0.235 | 0.135 | 0.119 | 0.167 | 0.140 | 0.160 | 9.000 | 9.000 | 9.000 | 9.000 | 9.000 | 9.000 |
| `falcon` | 0.412 | 0.427 | 0.459 | 0.457 | 0.392 | 0.438 | 10.000 | 10.000 | 10.000 | 10.000 | 10.000 | 10.000 |
| `vicuna` | 0.412 | 0.333 | 0.393 | 0.414 | 0.431 | 0.461 | 9.000 | 9.000 | 9.000 | 9.000 | 9.000 | 9.000 |
| `neural-chat` | 0.127 | 0.083 | 0.059 | 0.038 | 0.130 | 0.132 | 10.000 | 10.000 | 10.000 | 10.000 | 10.000 | 10.000 |
| `gemma` | 0.118 | 0.083 | 0.081 | 0.105 | 0.119 | 0.110 | 2.000 | 2.000 | 2.000 | 2.000 | 2.000 | 2.000 |
| `taxollama` | 0.490 | 0.354 | 0.348 | 0.405 | 0.365 | 0.352 | 9.208 | 9.167 | 9.000 | 9.152 | 9.123 | 9.100 |

| | $A$ | | | | | | r-conf | | | | | |
|---|---|---|---|---|---|---|---|---|---|---|---|---|
| | $bin0$ | $bin1$ | $bin2$ | $bin3$ | $bin4$ | $bin5$ | $bin0$ | $bin1$ | $bin2$ | $bin3$ | $bin4$ | $bin5$ |
| `gpt-3.5` | 0.314 | 0.396 | 0.511 | 0.476 | 0.558 | 0.534 | 8.167 | 8.290 | 8.010 | 7.840 | 7.948 | 7.935 |
| `llama-3` | 0.608 | 0.552 | 0.600 | 0.671 | 0.725 | 0.776 | 9.624 | 9.154 | 9.040 | 9.068 | 9.177 | 9.212 |
| `llama-2` | 0.490 | 0.448 | 0.444 | 0.429 | 0.444 | 0.475 | 8.000 | 8.000 | 8.000 | 8.000 | 8.000 | 8.000 |
| `mistral` | 0.235 | 0.219 | 0.185 | 0.152 | 0.185 | 0.174 | 9.000 | 9.000 | 9.000 | 9.000 | 9.000 | 9.000 |
| `falcon` | 0.402 | 0.438 | 0.400 | 0.448 | 0.418 | 0.438 | 10.000 | 10.000 | 10.000 | 10.000 | 10.000 | 10.000 |
| `vicuna` | 0.392 | 0.375 | 0.363 | 0.410 | 0.421 | 0.416 | 9.000 | 9.000 | 9.000 | 9.000 | 9.000 | 9.000 |
| `neural-chat` | 0.147 | 0.094 | 0.081 | 0.067 | 0.146 | 0.151 | 10.000 | 10.000 | 10.000 | 10.000 | 10.000 | 10.000 |
| `gemma` | 0.118 | 0.083 | 0.081 | 0.105 | 0.119 | 0.110 | 2.000 | 2.000 | 2.000 | 2.000 | 2.000 | 2.000 |
| `taxollama` | 0.696 | 0.510 | 0.496 | 0.500 | 0.479 | 0.521 | 9.210 | 9.278 | 9.167 | 9.219 | 9.237 | 9.167 |

Table 14: Accuracy and confidence over correct answers w.r.t. the HyperLex score bins on (top) HyperLex-FS and (bottom) Fellbaum-FS prompting.

Table 13, resp. Table 14, report a summary of the average accuracy $A$ and average confidence *r-conf* obtained by the models in the zero-shot, resp. few-shot, scenario for each of the HyperLex subsets corresponding to bins of the score interval, i.e., $[0.0, 1.0), [1.0, 2.0), \ldots, [5.0, 6.0]$, hereinafter referred to as $bin0, bin1, \ldots, bin5$. As it can be observed from both tables, there is no evident trend that relates the models' accuracy and confidence to the score bins. One exception would be represented by LLama3 and GPT-3: the former shows an average accuracy of around 0.36 in $bin0$ and $bin1$, before increasing in the subsequent bins for the zero-shot scenario, while the latter exhibits a similar behavior for the few-shot scenario, with an average accuracy below 0.4 in $bin0$ and $bin1$ for Fellbaum-FS, and around 0.32 in $bin0$ for HyperLex-FS, before increasing in the other bins.